\documentclass[conference]{IEEEtran}
\usepackage{times}
\usepackage{latexsym}

\usepackage{amsmath}
\usepackage{amssymb}
\usepackage{graphicx}
\usepackage{multirow}
\usepackage{tabularx}
\usepackage{enumitem}
\usepackage{hyperref}
\usepackage[numbers]{natbib} 

\usepackage{tablefootnote}
\usepackage[table]{xcolor}

\ifCLASSINFOpdf
\else
\fi
\begin{document}
%
\title{WikiContradiction: Detecting Self-Contradiction Articles on Wikipedia}

\author{\IEEEauthorblockN{Cheng Hsu$^{\dag}$, Cheng-Te Li$^{\dag}$, Diego Saez-Trumper$^{\ddag}$, Yi-Zhan Hsu$^{\dag}$}
\IEEEauthorblockA{$^{\dag}$Institute of Data Science, National Cheng Kung University, Taiwan\\
$^{\ddag}$Wikimedia Foundation Barcelona, Spain\\
Email: chengte@ncku.edu.tw, diego@wikimedia.org}
}


%


\maketitle

\begin{abstract}
While Wikipedia has been utilized for fact-checking and claim verification to debunk misinformation and disinformation, it is essential to either improve article quality and rule out noisy articles. Self-contradiction is one of the low-quality article types in Wikipedia. In this work, we propose a task of detecting self-contradiction articles in Wikipedia. Based on the ``self-contradictory'' template, we create a novel dataset for the self-contradiction detection task. Conventional contradiction detection focuses on comparing pairs of sentences or claims, but self-contradiction detection needs to further reason the semantics of an article and simultaneously learn the contradiction-aware comparison from all pairs of sentences. Therefore, we present the first model, Pairwise Contradiction Neural Network (PCNN), to not only effectively identify self-contradiction articles, but also highlight the most contradiction pairs of contradiction sentences. The main idea of PCNN is two-fold. First, to mitigate the effect of data scarcity on self-contradiction articles, we pre-train the module of pairwise contradiction learning using SNLI and MNLI benchmarks. Second, we select top-$K$ sentence pairs with the highest contradiction probability values and model their correlation to determine whether the corresponding article belongs to self-contradiction. Experiments conducted on the proposed WikiContradiction dataset exhibit that PCNN can generate promising performance and comprehensively highlight the sentence pairs the contradiction locates.
\end{abstract}


%
\IEEEpeerreviewmaketitle

\section{Introduction}
\label{sec:intro}
While social media brings convenient communication and allows easier interactions between people, it is also rooted in the dissemination of misinformation and disinformation \cite{fakedm17,fakeap20}. Misinformation refers to false and incorrect information while disinformation is purposefully manipulated news \cite{fakesuv20}. Wikipedia has been used as source for large-scale corpus of real claims and evidence documents \cite{odw19}, and thus has been adopted  for fact-checking and verification against fake and false information, such as WikiFactCheck-English \cite{afcw20}, FEVER \cite{fever18}, MultiFC \cite{multifc19}, WikiCitations \cite{cnwiki19}, and Wiki-Check \cite{trokhymovych2021wikicheck}. Wikipedia can be also utilized to construct knowledge graphs to enhance applications, such as recommender systems \cite{wikikg20}, question answering \cite{wikiqa20}, and dialogue Systems \cite{wikids19}. Although Wikipedia is widely exploited, dealing with noisy and low-quality Wikipedia articles is still critical. Therefore, is crucial for editors and fact-checkers to have some approaches to identify Wikipedia articles containing incorrect information.

To improve the quality of Wikipedia articles, this work dives into the detection of \textit{self-contradiction} articles in Wikipedia. An article is regarded as self-contradiction if it contains multiple claims or ideas that are inherently in disagreement. That said, if an article possesses at least two statements that contradict one another, we can say that this article contradicts itself, i.e., is self-contradiction. In Wikipedia, a ``Self-contradictory'' template\footnote{\url{https://en.wikipedia.org/wiki/Template:Self-contradictory}} is created to annotate self-contradiction articles. Editors can use the ``self-contradictory'' template to manually indicate whether an article is self-contradiction, resulting in the historical collection of self-contradiction articles\footnote{\url{https://en.wikipedia.org/wiki/Category:Self-contradictory_articles}}. We aim to accordingly create a dataset for detecting self-contradiction articles in Wikipedia. Besides, we further propose the first model, Pairwise Contradiction Neural Network (PCNN), for the detection task.

\begin{table}[!t]
\centering
\caption{Two examples of self-contradictory Wikipedia articles: ``\textit{Tyler Acord}''\tablefootnote{\url{https://en.wikipedia.org/w/index.php?title=Tyler_Acord&oldid=895550671}} and ``\textit{Pink Chanel suit of Jacqueline Bouvier Kennedy}''\tablefootnote{\url{https://en.wikipedia.org/w/index.php?title=Pink_Chanel_suit_of_Jacqueline_Bouvier_Kennedy&direction=prev&oldid=885388833}}. The texts that contradict with each other are highlighted in \textbf{bold} font. Contradicted parts tend to appear at different sentences within an article.
}
\label{tab:exarticles}
\begin{tabularx}{\linewidth}{l}
\hline
\rowcolor{gray!15}
\multicolumn{1}{c}{Wikipedia Article 1: \textit{Tyler Acord}} \\ \hline
\dots Tyler Acord (born September 12, 1990), better known by his stage \\ name Lophiile and formerly known as Scout, is an American record \\ producer, DJ, multi-instrumentalist, and songwriter \textbf{born in Lakewood,} \\ \textbf{Washington}. \dots Tyler Acord was \textbf{born in Renton, Washington} on \\ September 12, 1990. \dots
 \\ \hline\hline
\rowcolor{gray!15}
\multicolumn{1}{c}{Wikipedia Article 2: \textit{Pink Chanel suit of Jacqueline Bouvier Kennedy}} \\ \hline
\dots There was long a question among fashion historians and experts \\ whether the suit \textbf{was a genuine Chanel or a quality copy} purchased \\ from New York's Chez Ninon, a popular dress shop that imported \\ European label designs. \dots A number of sources claimed it \textbf{was more} \\ \textbf{than likely a copy of a Chanel} pink bouclé wool suit trimmed with a \\ navy blue collar \dots
 \\ \hline
\end{tabularx}%
\end{table}

Here we give two examples of self-contradictory articles in Wikipedia, as presented in Table~\ref{tab:exarticles}. For the first example, i.e., \textit{Tyler Acord}, it can be clearly found that two highlighted sentences are contradicted with one another because the birthplaces are different. For the second example, i.e., \textit{Pink Chanel suit of Jacqueline Bouvier Kennedy}, it is about the authenticity of the suit. One sentence mentions that the suit could be a genuine or a copy, and the other emphasizes a copy.

Detecting self-contradiction articles is different from conventional contradiction detections in various text data. The input of typical contradiction detection is a pair of sentences or claims, and its goal is to classify whether they contradict each other \cite{fct08acl}. Previous work has explored contracdictions in domains such as scientific reviews \cite{adcon15,pconi18}, short-text posts on social media \cite{msmcon16,cafcon18}, ``rumorous claims'' \cite{condrc16}, and commercial item reviews \cite{condwe17,rcodan19}. As for the self-contradiction detection proposed in this work, we are given an article containing a number of sentences, and the task is to simultaneously classify the article as self-contradiction or not and identify which pairs of sentences are contradicted with one another. The key difference is that detecting self-contradiction requires a model to understand the semantics and topics of the input article, in addition to have pairwise comparisons between sentences, so that the sentences whose meanings contradict with other sentences or the whole article can be highlighted. Although an existing study \cite{idupc15} found that few of near-duplicate sentences can contradict with each other in Wikipedia. They simply perform sentence clustering with lexical Jaccard similarity, but do not address the problem of finding contradiction sentences with different phrasings.

In this paper, we create a new dataset, WikiContradiction, which contains both self-contradiction and non-contradiction articles in Wikipedia. We develop the first model, Pairwise Contradiction Neural Network (PCNN), for the detection task. The main idea of PCNN is three-fold. First, we fine-tune the Sentence-BERT model to generate the representations of sentences in a Wikipedia article. Second, we pre-train a pairwise contradiction learning network to generate the contradiction probability of each pair of sentences. Third, we select top-$K$ sentences with the highest contradiction probabilities, and utilize their embeddings to generate the binary classification outcomes.

\begin{table*}[!t]
\centering
\caption{Comparison of relevant studies on contradiction detection. Note that column ``Type'' means contradiction type (Pairwise vs. self contradiction, and column ``Explainability'' indicates whether the method can highlight which parts of the model input contradict with each other.}
\label{tab:relatedtab}
\resizebox{0.925\textwidth}{!}{%
\begin{tabular}{c|c|c|c|c|c|c}
\hline
 & Data & Type & Model Input & Feature Extraction & Classifier & Explainability \\ \hline\hline
\cite{adcon15} & Scientific Claims & Pair & Sentence Pair & Linguistic & SVM &  \\ \hline
\cite{pconi18} & Item Reviews & Pair & Sentence Pair & Aspects \& Sentiments & Decision Tree &  \\ \hline
\cite{msmcon16} & Event Tweets & Pair & Sentence Pair & POS Tags \& Dependency  Parsing & MaxEnt &  \\ \hline
\cite{cafcon18} & Product Tweets & Pair & Sentence Pair & Sentiments & CBOW \& ESIM &  \\ \hline
\cite{condrc16} & Rumorous Claims & Pair & Sentence Pair & Text Similarity & Random Forest &  \\ \hline
\cite{condwe17} & Video Snippets & Pair & Sentence Pair & Word Embeddings & CNN &  \\ \hline
\cite{rcodan19} & Video Snippets & Pair & Sentence Pair & Word Embeddings & GRU &  \\ \hline
\textbf{This work} & \textbf{Wikipedia Articles} & \textbf{Self} & \textbf{Article} & \textbf{Pre-Training \& Fine-Tuning} & \textbf{MLP} & \textbf{\checkmark} \\ \hline
\end{tabular}%
}
\end{table*}

Experiments conducted on our WikiContradiction dataset deliver three main findings. First, PCNN can apparently outperform typical document classification models. Second, the pre-training of pairwise contradiction learning has the most significant contribution to the detection performance. Third, the conducted case studies exhibit that PCNN can truly identify the most contradictory pairs of sentences regarding contradictions within an article.

Below we list the contributions of this work.
\begin{itemize}
\item We create a novel wiki dataset, WikiContradiction, for self-contradiction Wikipedia article detection\footnote{Data and code can be access at this link: \url{https://github.com/Wiki-Contradictory/Wiki-Self-Contradictory/}}. To the best of our knowledge, it is the first dataset for the self-contradiction detection task on Wikipedia.
\item We define the task of detecting self-contradiction articles and highlighting contradiction sentence pairs, and solve it by developing a novel model, Pairwise Contradiction Neural Network (PCNN). We propose to pre-train PCNN using two benchmarks SNLI and MNLI, and fine-tune it via our WikiContradiction dataset.
\item Experimental results show that PCNN can not only lead to the promising performance in both imbalanced and balanced settings, but also highlight the most contradictory pairs of contradicting sentences in an article.
\end{itemize}

This paper is organized as follows. We review the relevant related work in Section~\ref{sec:related}, next in Section~\ref{sec:data} we describe  WikiContradiction dataset. Section~\ref{sec:model} gives the technical details of the proposed PCNN model. We describe the evaluation settings and experimental results in Section~\ref{sec:exp}. Last, we discuss the implication and limitation in Section~\ref{sec:discuss}, and conclude this work in Section~\ref{sec:conc}.







\section{Related Work}
\label{sec:related}

\textbf{Contradiction Detection.}
The typical contradiction detection task aims to classifying whether one sentence or claim is contradicted with another. The input instance is a pair of sentences and claims. A variety of supervised learning-based models has been developed. Alamri and Stevensony~\cite{adcon15} extract linguistic features, and use support vector machine to classify potentially contradictory scientific claims. Ismail et al.~\cite{pconi18} take advantage of review polarity to train supervised classifiers for predicting contradiction intensity of item reviews. Lendvai et al.~\cite{msmcon16} rely on part-of-speech and dependency parsing, together with the MaxEnt Classifier~\cite{maxent}, to classify the contradiction of social media posts. Li et al.~\cite{cafcon18} incorporate sentiment analysis with contradiction detection on Amazon's customer reviews. Lendvai and Reichel~\cite{condrc16} utilize a set of textual similarity features, including vocabulary overlap, local alignment, corpus statistics, and apply random forest to classify whether a pair of rumorous claims is contradictory. Li et al.~\cite{condwe17} learn contradiction-specific word embeddings, and use convolutional neural network to recognize contradiction relations between a pair of sentences. Besides, Tan et al.~\cite{rcodan19} propose a dual attention-based gated recurrent unit to learn aspect-level sentiments for recognizing conflict opinions. All of these approaches to contradiction detection compare two sentences or claims from various aspects and side information. Nevertheless, they cannot be directly applied to the self-contradiction detection of an article that involves all pairwise reasoning.

In Table~\ref{tab:relatedtab} we compare our work with existing studies on contradiction detection. There are six different aspects being compared, including data for contradiction detection (``Data''), contradiction type (``Type''), model input, feature extraction, classifier, and explainability. Based on that table, to best of our knowledge we are the first to deal with self-contradiction detection within an article while all of past work focus on pairwise contradiction detection for a given sentence pair. In addition, in extracting features, we do not rely on hand-crafted engineering, but utilize pre-training and fine-tuning strategies to learn features regarding contradiction. Furthermore, our model is the only one that can provide model explainability through highlighting which parts of the model input contradict with each other.

\textbf{Recognizing Textual Entailment.}
The task of recognizing textual entailment aims to predict whether a given textual premise entails or implies a given hypothesis, i.e., a binary classification task~\cite{rtesuv13}. 
One of the essential entailing relations between sentences is a contradiction. Early approaches rely on the sentence structure with linguistic resources, like WordNet~\cite{wordnett} and Verbnet~\cite{werbnett}, to bring semantics into the estimation of entailment by alignment~\cite{dncrte08} and transformation~\cite{trmrte17}. The supervised Context-Enriched Neural Network (CENN)~\cite{cenn17} utilize multiple embeddings from different contexts to better represent text pairs, along with an attention mechanism to combine them for predicting the entailment relation. Silva et al.~\cite{tekg18} further explore structured knowledge graphs to better classify and explain the entailment relation. Recently, with the advances of techniques for Natural Language Inference (NLI), a category of textual entailment is to classify a pair of text pieces as \textit{entailment}, \textit{neutral} or \textit{contradiction}. A number of methods are now based on deep learning~\cite{lstmnli17,gong18}, and especially resorting to attention mechanisms~\cite{datt16,entatt16,tfm17}. Well-known transformer-based models, such as BERT~\cite{bert19}, RoBERTa~\cite{roberta19}, and XLNet~\cite{xlnet19}, further lead to state-of-the-art performances. 

\textbf{Document Classification.}
The detection of self-contradiction articles can be considered as the document classification task because it performs binary classification based on the whole textual content of a document. Conventional approaches extract features like bag-of-words, TFIDF, n-gram, word embeddings~\cite{w2v13}, and Doc2Vec~\cite{d2v14}, and apply multinomial logistic regression and support vector machine for document classification~\cite{docclf12,textcls19}. Techniques on deep learning can generate better representations of documents, and thus significantly improve the performance of document classification~\cite{docclssuv}. Typical methods include TextCNN~\cite{textcnn14}, CharCNN~\cite{charcnn15}, and hierarchical attention networks (HAN)~\cite{han16}, and RegLSTM~\cite{lstm97,reglstm19}. Recently, the pre-training language representation learning models~\cite{bert19,roberta19} had been adopted for document classification with promising performance.


\section{WikiContradiction Dataset}
\label{sec:data}

\begin{figure}[!t]
\centering
\includegraphics[width=0.65\linewidth]{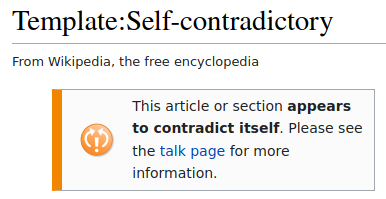}
\caption{The self-contradictory template is used by editors to signal articles containing self-contradiction information. It is rendered at the top of the article or in the section that contains the self-contradiction.}
\label{fig:template}
\end{figure}

Wikipedia editors utilize the ``Self-contradictory Template'' to tag articles that contain contradictory information. ``Templates are pages that are embedded into other pages to allow for the repetition of information''.~\footnote{\url{https://en.wikipedia.org/wiki/Wikipedia:Templates}} Templates can be used - among other things - to signal problems with articles' content, allowing readers and other editors to understand issues with an article or a specific piece of content. One example of a well-known template is the ``citation needed'' tag~\cite{cnwiki19}.
We can assume that templates used to signal problematic content had a good precision, because adding a template requires expertise on Wikipedia, meaning that the editor adding that template is likely to have a good knowledge of how Wikipedia works. In that sense, we can consider this as a high-quality manual annotation of data. This methodology has already been used to generate high-quality datasets to indicate other content reliability issues in Wikipedia~\cite{wong2021wiki}. However, for the same reasons - requiring expert editors - the recall could be low, because we cannot assume that every article has been reviewed by an expert editor. In our dataset, the 87\% of the users that added a template had over 1,000 edits on Wikipedia at the time they added the self-contradictory template. The usage of templates is not trivial and requires knowledge of the MediaWiki functionalities. Moreover, templates such as ``self-contradictory'' are not very popular (appears in less than 1\% of English Wikipedia articles), meaning that users adding such templates have a deep knowledge on Wikipedia's conventions. Therefore, we consider Wikipedia editors with more than 1,000 edits as high-quality annotators.

\begin{figure}[!t]
\centering
\includegraphics[width=1.0\linewidth]{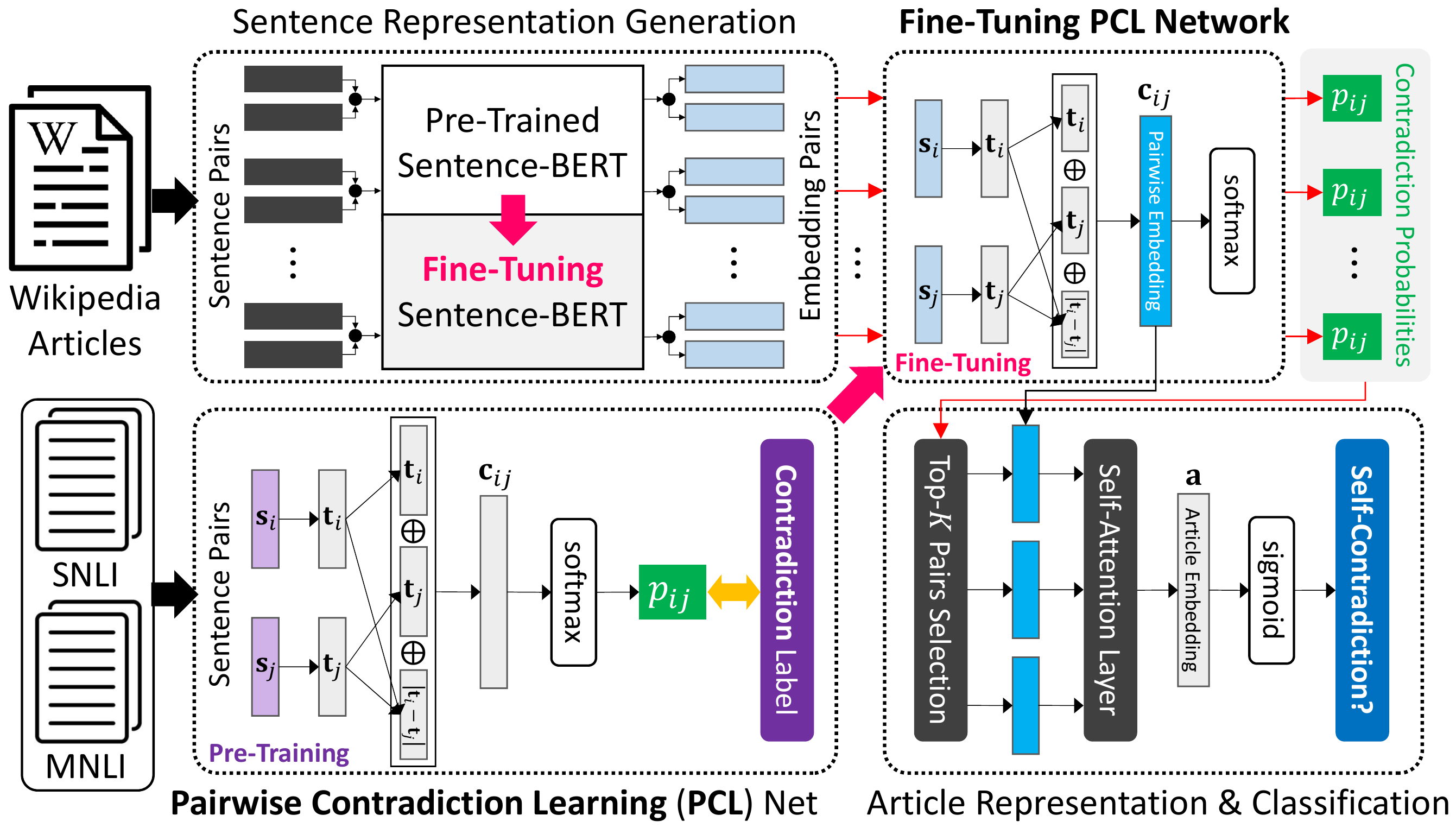}
\caption{Model architecture of the proposed PCNN.}
\label{fig:pcnnflow}
\end{figure}

To build a balanced dataset with examples of self-contradiction and non-self-contradiction (normal) articles, we look at all the versions (a.k.a. revisions) of all articles in English Wikipedia, and select the ones that have had the self-contradiction added in one old version. Next, we search for newer versions where that template has been removed, meaning that editors have reviewed the new version of the article, and removed the template showing that the self-contradiction problem has been resolved in this new version. This methodology allows us to build a balanced dataset, where both categories has been annotated by experts. 

We took all articles in English Wikipedia until March 2020, and running a simple string matching process we detected all the revisions that contain a self-contradiction template. Next, we scan all the newer versions of those articles until finding a version of the article that does not contain that template. In total, we find $2321$ articles with the template, where $1105$ had a version that resolved the self-contradiction. We discarded $1216$ articles that do no resolve the contradiction. That said, eventually we have $1105$ positive (self-contradiction) articles and $1105$ negative (non-self-contradiction) articles. When splitting the data into training and testing, we ensure both self-contradiction and non-self-contradiction versions of an article totally appear in either training or testing set to avoid the leaking of classification label.

\section{The Proposed Model: PCNN}
\label{sec:model}

Let $a$ be a Wikipedia article, consisting of $n$ sentences $\{s_i\}_{i=1}^n$. We treat the Wikipedia article self-contradiction detection task as the binary classification problem. Specifically, each article $a$ can be true (i.e., self-contradiction, $y_a=1$) or false (i.e., non-contradiction, $y_a=0$). For a self-contradiction Wikipedia article $a$, there exists at least two sentences $s_i$ and $s_j$ whose semantic meanings or referring facts contradict with one another. In addition to the detection task, we also aim to learn a ranking list $L$ from all pairs of sentences in $\{s_i\}_{i=1}^n$, according to the prediction probabilities, where $L_k$ denotes the $k$-th most explainable pair of sentences that contradict with each other. 

\textbf{Problem Definition: Wikipedia Self-Contradiction Article Detection.} Give an article $a$ in Wikipedia, our task is to learn a self-contradiction detection function $f: f(a)\rightarrow(\hat{y},L)$, such that it maximizes the classification probability with explainable pairs of sentences ranked highest in list $L$.

\textbf{PCNN Model Architecture.} The architecture of PCNN model is presented in Figure~\ref{fig:pcnnflow}, which consists of four components. The first is \textit{Sentence Representation Generation}. Given a Wikipedia article, we generate the representation vector for each sentence through a pre-trained Sentence-BERT model. This component with Sentence-BERT network architecture is fine-tuned based on our data. The second is \textit{Pre-Training Pairwise Contradiction Learning (PCL) network} based on two natural language inference benchmarks SNLI and MNLI. PCL aims at generating a \textit{contradiction probability} value that depicts how two sentences contradict with each other. The PCL pre-training is treated as initializing PCL model parameters that capture comparative semantics between sentences. The third is \textit{Fine-Tuning PCL Network}, which is end-to-end trained with fine-tuning Sentence-BERT component and followed by the last component. We fine-tune the PCL network and produce the contradiction probability of every sentence pair in our data. The last component is \textit{Article Representation \& Classification}. We select the most ``suspicious'' sentence pairs based on their contradiction probabilities, accordingly utilize a self-attention layer to encode their correlation, and generate the final binary classification outcome.

\subsection{Sentence Representation}
Since self-contradiction detection involves the inspection between sentences within an article, we can consider the task as sentence-pair regression tasks like semantic textual similarity mentioned in BERT~\cite{bert19} and RoBERTa~\cite{roberta19}. Sentence-BERT~\cite{sbert19} improves the representation capability for sentences through siamese and triplet networks, in which BERT networks are fine-tuned with shared weights. In order to have sentences being semantically comparable with an article, we fine-tune the pre-trained Sentence-BERT model\footnote{\url{https://www.sbert.net/}} to generate sentence embeddings, denoted as $\mathbf{s}_i$ for each sentence $s_i$. The derived sentence embeddings are used as the initial vectors of our pairwise sentence contradiction learning.

\subsection{Pairwise Contradiction Learning}
In detecting self-contradiction articles, we compare the semantics of sentences and generate a contradiction probability depicting the degree that two sentences contradict each other. Intuitively we should enumerate all pairs of sentences within an \textit{article} for semantically-contradiction comparison. Instead, we learn to produce the contradiction probability of each sentence pair within a \textit{paragraph}. The reason is two-fold. First, existing studies have pointed out that sentences within a paragraph of an article in Wikipedia are semantically coherent and topically consistent~\cite{wikinlp09,wikiwrite16,wikisum18}. Such a discovery encourages us to first examine sentences in a paragraph in a pairwise manner. We will verify whether paragraph-level pairwise sentence comparison is better than the article-level version in the experiment. Second, popular articles in Wikipedia can contain hundreds of sentences. Enumerating all pairs of sentences can bring high computational cost.  

Given the embeddings of two sentences, $\mathbf{s}_i$ and $\mathbf{s}_j$, we pre-train a pairwise contradiction learning (PCL) model that generates pairwise sentence embedding $\mathbf{c}_{ij}$ and a probability $p_{ij}$. We first utilize a learnable weight matrix $\mathbf{W}_t$ to generate intermediate vectors, given by: $\mathbf{t}_i = \mathbf{W}^t\mathbf{s}_i$ and $\mathbf{t}_j = \mathbf{W}^t\mathbf{s}_j$. Then we concatenate the vectors $\mathbf{t}_i$ and $\mathbf{t}_i$ with the element-wise difference $|\mathbf{t}_i-\mathbf{t}_j|$, and multiply it with a trainable weight $\mathbf{W}^r$, along with the softmax function, to generate the contradiction probability between sentences $s_i$ and $s_j$, denoted as $p_{ij}$, given by:
\begin{equation}
\begin{split}
\mathbf{c}_{ij} &= \left(\mathbf{t}_i \parallel \mathbf{t}_j \parallel |\mathbf{t}_i-\mathbf{t}_j|\right),\\
p_{ij} &= \sigma\left(\mathbf{W}^r\mathbf{c}_{ij}\right)
\end{split}
\end{equation}
where $\sigma$ is the softmax function, $\parallel$ denotes the concatenation operation, and $\mathbf{W}^r\in\mathbb{R}^{3d\times 2}$. 
The cross-entropy loss is employed for optimization. 
Here the pre-training is performed using the set of sentences with ``contradiction'' label (i.e., binary classification) in both Stanford Natural Language Inference dataset (SNLI)\footnote{\url{http://nlp.stanford.edu/projects/snli/}}
~\cite{snli15} 
and the Multi-Genre NLI (MNLI) dataset\footnote{\url{https://cims.nyu.edu/~sbowman/multinli/}}~\cite{mnli18}. 
Eventually we fine-tune the pre-trained model to generate pairwise sentence embedding $\mathbf{c}_{ij}$ and the contradiction probability $p_{ij}$ using our complied Wiki-Contradiction data for the detection of self-contradiction articles.

\subsection{Article Representation \& Classification}
We aim at generating the representation of the given article, and accordingly produce its self-contradiction probability. Since the number of sentences involving self-contradiction tends to be limited in an article, we consider only the most contradicting sentence pairs to determine whether an article is self-contradiction. Sentence pairs with higher probability $p_{ij}$ are utilized to learn the article representation. In other words, we select the embedding vectors $\{\mathbf{c'}^{k}_{ij}\}_{k=1}^{K}$ of sentence pairs with top-$K$ probabilities. 

Sentence pairs in an article can be correlated with one another. The semantics (e.g., coherent or not in their meanings) of contradiction pairs can differ from that of non-contradiction sentence pairs. Therefore, we need to model how different sentence pairs correlate with each other and how they contribute to self-contradiction detection in article representation learning. We exploit a self-attention layer~\cite{tfm17} to achieve this goal. Specifically, we can generate the article embedding $\mathbf{a}$ based on:
\begin{equation}
\mathbf{a} = \frac{\sum_{k=1}^{K}\alpha_{k}\mathbf{w}^{s}\mathbf{c'}^{k}_{ij}}{K},
\end{equation}
where $\alpha_{k}$ is the attention weight vector (estimating the contribution of each sentence pair) obtained from a softmax function, which is applied to the dot product of two different transformations on $\mathbf{c'}^{k}_{ij}$, and $\mathbf{w}^{s}$ is the learnable parameter.

To produce the classification probability, we feed the article embedding $\mathbf{a}$ into a one-hidden-layer feed-forward network, together with a sigmoid function, to generate the probability of classifying as self-contradiction. The cross-entropy loss is used for model optimization. In detail, we use a batch-size of $16$, Adam optimizer~\cite{adam} with a learning rate $2e-5$, and a linear learning rate warm-up over 10\% of the training data.

\section{Experiments}
\label{sec:exp}

\subsection{Evaluation Settings}
Although the compiled dataset with self-contradiction and non-self-contradiction Wikipedia articles are balanced, it is relatively rare to have self-contradiction ones in the real world.
Hence, we divide the experiments into two sets, \textit{balanced} and \textit{imbalanced}. For the balanced setting, we randomly sample $10$ sets of the equal article number for each class.
For the imbalanced settings, we change the ratio of positive (self-contradiction) and negative (non-self-contradiction) articles, i.e., 10\%:90\%, 30\%:70\%, and 50\%:50\%, together with TR=80\%, and randomly generate $10$ sets for each imbalanced ratio. 
We use Precision (Pre), Recall (Rec), F1, and Accuracy (Acc) as the evaluation metrics. The default training-test split is 80\%:20\%. We report the average results. We will also vary the training percentages (TR) with 20\%, 40\%, 60\%, and 80\%. 

We compare the proposed PCNN with four baselines. (a) \textbf{Random}: determining the self-contradiction or not in a random manner; (b) \textbf{LSTM}~\cite{lstm97}: sequentially feeding the FastText Wiki pre-trained word embeddings\footnote{\url{https://fasttext.cc/docs/en/pretrained-vectors.html}} into LSTM, and utilizing the last hidden layer to produce the binary classification; (c) \textbf{HAN}~\cite{han16}: a well-known deep model with hierarchical attention networks for document classification; and (d) \textbf{BERT}~\cite{bert19}: fine-tuning the BERT model to the training data. The hyperparameters of competing methods are set by following the settings mentioned in respective studies, and their word embedding dim is $32$, and all of their intermediate embedding dim are $64$.

All experiments are conducted with PyTorch running on GPU machines (Nvidia GeForce GTX 1080 Ti). The default settings for the hyperparameters of PCNN is listed here. Both the dimensionalities of sentence embedding and pairwise sentence embedding are $128$. The dimensionality of article embedding is $64$. The selection of top-$K$ most contradiction sentence pairs is set with $K=10$.

\begin{table}[!t]
\centering
\caption{Main results under various training ratios (TR) under the balanced setting. Method with the highest scores is highlighted in \textbf{bold}.}
\label{tab:mainres}
\resizebox{0.95\linewidth}{!}{%
\begin{tabular}{c|l|c|c|c|c}
\hline
 &  & Pre & Rec & F1 & Acc \\ \hline
\multirow{5}{*}{TR=20\%} & Random & 0.4977 & 0.4888 & 0.4932 & 0.4932 \\ \cline{2-6} 
 & LSTM & 0.5026 & 0.5475 & 0.5238 & 0.5022 \\ \cline{2-6} 
 & BERT & 0.5201 & 0.5837 & 0.5501 & 0.5226 \\ \cline{2-6} 
 & HAN & 0.5361 & 0.5701 & 0.5526 & 0.5362 \\ \cline{2-6} 
 & PCNN & \textbf{0.5637} & \textbf{0.5903} & \textbf{0.5767} & \textbf{0.5608} \\ \hline\hline
 &  & Pre & Rec & F1 & Acc \\ \hline
\multirow{5}{*}{TR=40\%} & Random & 0.5024 & 0.4570 & 0.4786 & 0.5022 \\ \cline{2-6} 
 & LSTM & 0.4939 & 0.5565 & 0.5234 & 0.4932 \\ \cline{2-6} 
 & BERT & 0.4897 & 0.5284 & 0.5129 & 0.4886 \\ \cline{2-6} 
 & HAN & 0.5191 & 0.5524 & 0.5359 & 0.5203 \\ \cline{2-6} 
 & PCNN & \textbf{0.6026} & \textbf{0.6108} & \textbf{0.6067} & \textbf{0.6058} \\ \hline\hline
 &  & Pre & Rec & F1 & Acc \\ \hline
\multirow{5}{*}{TR=60\%} & Random & 0.5196 & 0.4796 & 0.4988 & 0.5128 \\ \cline{2-6} 
 & LSTM & 0.5041 & 0.5520 & 0.5269 & 0.5045 \\ \cline{2-6} 
 & BERT & 0.5152 & 0.5339 & 0.5244 & 0.5158 \\ \cline{2-6} 
 & HAN & 0.5551 & 0.6080 & 0.5936 & 0.5633 \\ \cline{2-6} 
 & PCNN & \textbf{0.6071} & \textbf{0.6153} & \textbf{0.6112} & \textbf{0.6085} \\ \hline\hline
 &  & Pre & Rec & F1 & Acc \\ \hline
\multirow{5}{*}{TR=80\%} & Random & 0.5167 & 0.4886 & 0.5023 & 0.5158 \\ \cline{2-6} 
 & LSTM & 0.5254 & 0.5610 & 0.5426 & 0.5271 \\ \cline{2-6} 
 & BERT & 0.5307 & 0.5339 & 0.5164 & 0.5651 \\ \cline{2-6} 
 & HAN & 0.5840 & 0.6169 & 0.6056 & 0.5803 \\ \cline{2-6} 
 & PCNN & \textbf{0.6318} & \textbf{0.6289} & \textbf{0.6303} & \textbf{0.6312} \\ \hline
\end{tabular}%
}
\end{table}


\begin{table}[!t]
\centering
\caption{Performance comparison on top-$k$ predicted self-contradiction articles.}
\label{tab:resk}
\resizebox{0.7\linewidth}{!}{%
\begin{tabular}{c|c|c|c}
\hline
\textit{Precision@} & $k=10$ & $k=20$ & $k=30$ \\ \hline
LSTM & 0.5145 & 0.5503 & 0.5264 \\ \hline
BERT & 0.5361 & 0.6071 & 0.5383 \\ \hline
HAN & 0.6032 & 0.6256 & 0.6350 \\ \hline
PCNN & \textbf{0.6508} & \textbf{0.6812} & \textbf{0.6616} \\ \hline\hline
\textit{Recall@} & $k=10$ & $k=20$ & $k=30$ \\ \hline
LSTM & 0.0113 & 0.0249 & 0.0339 \\ \hline
BERT & 0.0123 & 0.0271 & 0.0363 \\ \hline
HAN & 0.0136 & 0.0309 & 0.0392 \\ \hline
PCNN & \textbf{0.0158} & \textbf{0.0362} & \textbf{0.0459} \\ \hline
\end{tabular}%
}
\end{table}

\subsection{Experimental Results}
\textbf{Main Results.}
The performance comparison of PCNN with the competing methods under the balanced setting is reported in Table~\ref{tab:mainres}, in which the performance scores are derived based on all testing sentences. We can find that PCNN consistently outperforms the other methods over different training percentages and various evaluation metrics. Such results exhibit the promising capability of detecting self-contradiction articles by PCNN. The reason is two-fold. First, PCNN models the pairwise semantics between sentences that can compare whether there exists a conflict, while document classification models HAN and LSTM simply learn the whole article meaning. Second, fine-tuning the pre-training language model BERT has some effect, but is not that good as PCNN. Such an outcome could result from that all sentences and their pairs are equally treated and learned in BERT, but the contradiction tends to appear in a few of sentence pairs that need to be emphasized by the model. Although PCNN leads to better results by tackling these issues, the scores of PCNN with 80\% trained data are still not higher than $0.65$. This tells that detecting self-contradiction articles is indeed challenging.

\textbf{Ranking Results.}
By using the prediction probabilities reported by a model, we can have a ranking evaluation, in addition to the overall performance shown in Table~\ref{tab:mainres}. To be specific, we generate the scores of Precision@$k$ and Recall@$k$ with $k=\{10,20,30\}$ based on $k$ articles with the highest prediction probabilities. The results are displayed in Table~\ref{tab:resk}. We can observe that PCNN again leads to the highest scores on Precision and Recall. Although the Recall scores are quite low, the superiority of PCNN can be still maintained. In other words, PCNN can effectively find the self-contradiction articles by presenting top-$k$ ones.

\begin{figure}[!t]
\centering
\includegraphics[width=0.9\linewidth]{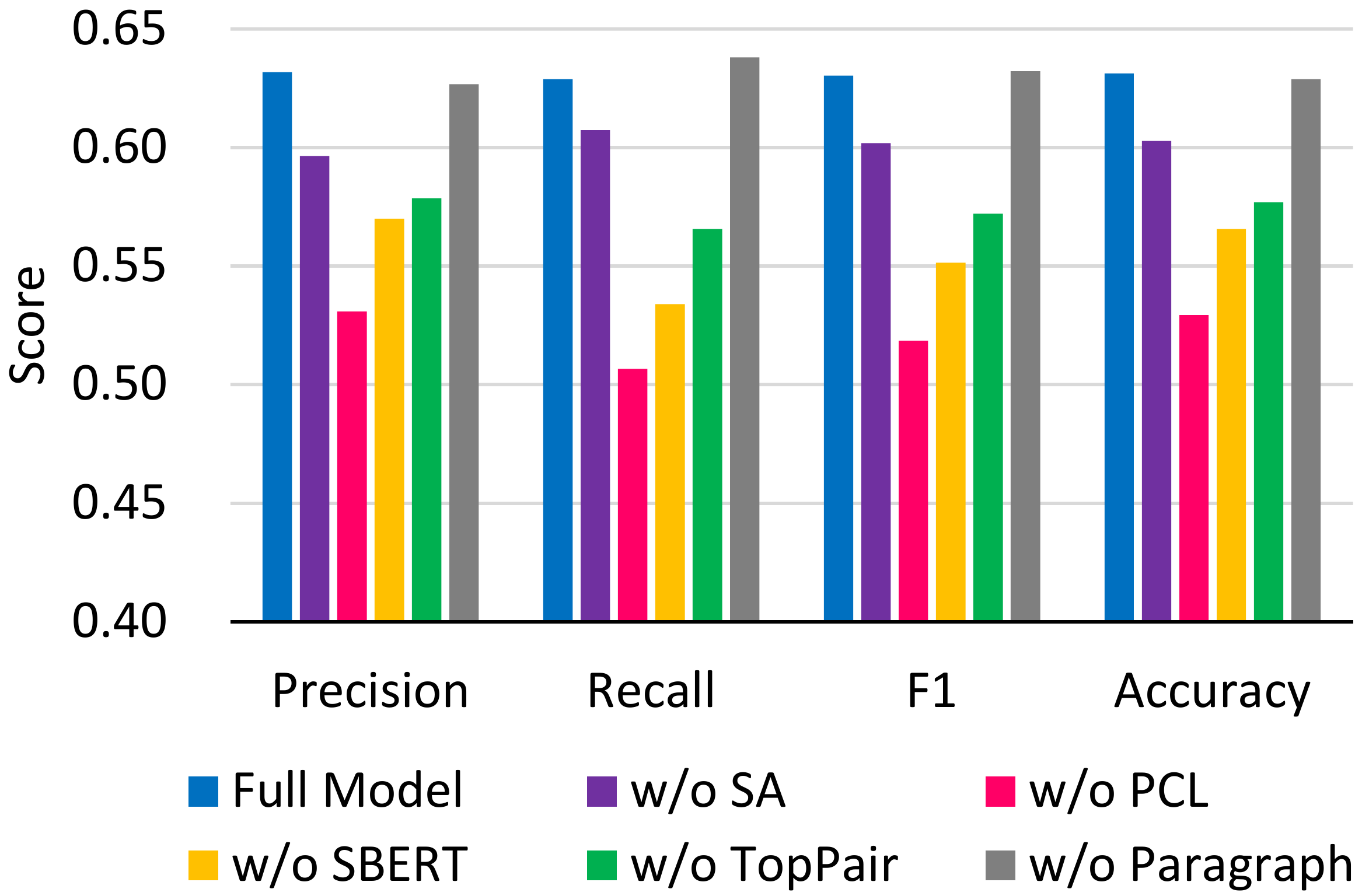}
\caption{Ablation study for the proposed PCNN.}
\label{fig:ablation}
\end{figure}

\textbf{Ablation Study.}
We conduct an ablation study to examine whether every component in PCNN does take effect. By removing each component from the full PCNN model (\textbf{Full Model}), we report the performance in terms of four metrics. There are five components to be investigated in this ablation study, as listed below. 
\begin{itemize}
\item Full model without the self-attention layer (\textbf{w/o SA}): utilizing simple concatenation to fuse all selected sentence pairs' embeddings, rather than the self-attention layer. 
\item Full model without pairwise contradiction learning (\textbf{w/o PCL}): replacing pairwise contradiction learning with simply concatenating the embeddings of two sentences, i.e., no contradiction comparison between sentences. 
\item Full model without Sentence-BERT (\textbf{w/o SBERT}): using LSTM to replace Sentence-BERT in learning sentence representations; 
\item Full model without the selection of top-$K$ contradiction sentence pairs (\textbf{w/o TopPair}): employing all sentence pairs for self-attention and prediction, no filtering out less contradiction ones; 
\item Full model without considering each paragraph for PCL (\textbf{w/o Paragraph}: utilizing all sentence pairs for PCL, instead of applying paragraph-level PCL.
\end{itemize}

The performance comparison on the ablation study is presented in Figure~\ref{fig:ablation}. The results bring several findings. First, every component, except for the paragraph-level PCL, in the proposed PCNN does contribute to the performance. Nevertheless, dividing an article into several paragraphs for PCL can be still to lower down the computational cost, i.e., improving the learning time efficiency, since we can perform PCL in parallel over paragraphs. Second, among all components, PCL brings the most significant performance improvement. This result verifies that the semantic and contradiction comparison between sentences can effectively capture the conflicts. Third, the components of Sentence-BERT and selecting top-$K$ sentence pairs are also useful. These exhibit the importance of sentence representation learning and eliminating non-conflicting sentence pairs. We would believe future advances on sentence representation learning and fine-grained determination of top-$K$ can further improve the performance. Last, the self-attention layer has a relatively minor contribution. Such an outcome informs us that the semantic modeling between ``sentences'' is more influential than that between ``sentence pairs.'' In summary, the results of the ablation study provide guidance on where future extensions can be performed to improve the detection of self-contradiction articles.

\begin{figure}[!t]
\centering
\includegraphics[width=1.0\linewidth]{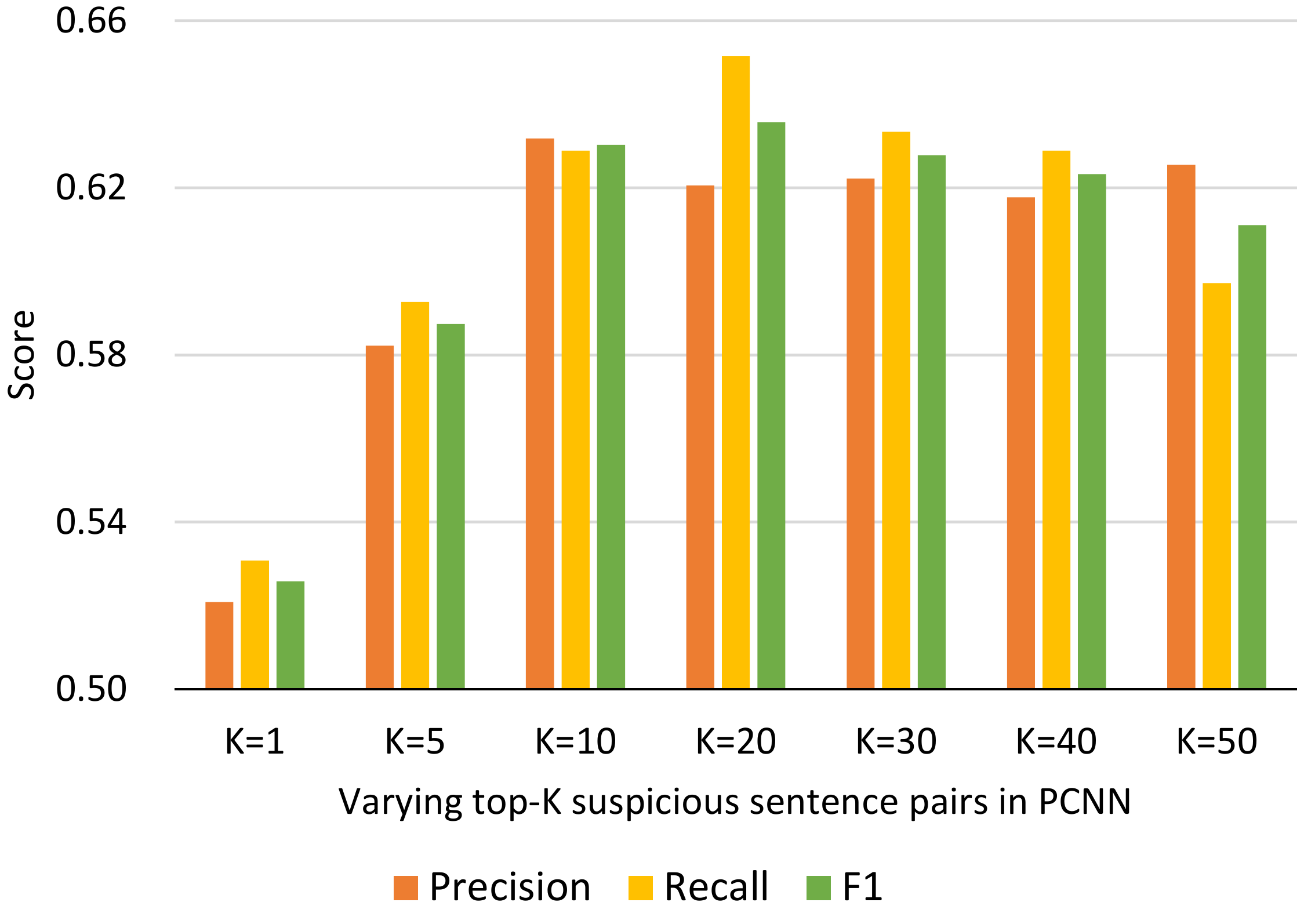}
\caption{Effect on varying top-$K$ pairs in PCNN.}
\label{fig:topkpair}
\end{figure}

\textbf{Top-$K$ Sentence Pairs.}
Our PCNN requires the selection of top-$K$ contradiction sentence pairs to classify the self-contradiction of an article. We aim at examining how the number $K$ of sentence pairs affects the performance. By varying $K$ numbers in PCNN, $K=\{1,5,10,20,30,40,50\}$, the results on Precision, Recall, and F1 are shown in Figure~\ref{fig:topkpair}. Based on the resultant scores, PCNN can generate better performance when $K$ is around $10$ to $20$. This implies that a moderate selection of contradiction sentence pairs can benefit the performance. A very limited $K$ number (e.g., $K=1$ or $K=5$) cannot collect all of the evidences of contradiction between sentences. Selecting too many sentence pairs (e.g., $K=50$ or $K=40$) can include non-contradiction sentence pairs, and thus damages the performance. We think a moderate number $K$ is reasonable since it is less possible for a sentence to contradict too many other sentences in an article. 

\begin{table}[!t]
\centering
\caption{Performance comparison by sampling different ratios of positive (\textbf{P}, i.e., self-contradiction) and negative (\textbf{N}, i.e., non-self-contradiction) articles.}
\label{tab:imbalanced}
\resizebox{0.9\linewidth}{!}{%
\begin{tabular}{c|c|c|c|c}
\hline
 & \textbf{P:N} & 10\%:90\% & 30\%:70\% & 50\%:50\% \\ \hline
\multirow{5}{*}{Precision} & Random & 0.0829 & 0.3201 & 0.5167 \\ \cline{2-5} 
 & LSTM & 0.1091 & 0.3715 & 0.5254 \\ \cline{2-5} 
 & BERT & 0.0990 & 0.3607 & 0.5074 \\ \cline{2-5} 
 & HAN & 0.1207 & 0.4027 & 0.5840 \\ \cline{2-5} 
 & PCNN & \textbf{0.1414} & \textbf{0.4298} & \textbf{0.6318} \\ \hline\hline
 & \textbf{P:N} & 10\%:90\% & 30\%:70\% & 50\%:50\% \\ \hline
\multirow{5}{*}{Recall} & Random & 0.4090 & 0.4965 & 0.4886 \\ \cline{2-5} 
 & LSTM & 0.5681 & 0.5510 & 0.5610 \\ \cline{2-5} 
 & BERT & 0.5198 & 0.5374 & 0.5339 \\ \cline{2-5} 
 & HAN & 0.5682 & 0.6054 & 0.6289 \\ \cline{2-5} 
 & PCNN & \textbf{0.6191} & \textbf{0.6462} & \textbf{0.6389} \\ \hline\hline
 & \textbf{P:N} & 10\%:90\% & 30\%:70\% & 50\%:50\% \\ \hline
\multirow{5}{*}{F1} & Random & 0.1379 & 0.3833 & 0.5023 \\ \cline{2-5} 
 & LSTM & 0.1831 & 0.4438 & 0.5426 \\ \cline{2-5} 
 & BERT & 0.1654 & 0.4316 & 0.5164 \\ \cline{2-5} 
 & HAN & 0.1992 & 0.4836 & 0.6056 \\ \cline{2-5} 
 & PCNN & \textbf{0.2329} & \textbf{0.5163} & \textbf{0.6303} \\ \hline
\end{tabular}%
}
\end{table}

\begin{table*}[!t]
\centering
\caption{Results on case studies, in which ``GT'' means the ground-truth, and ``Pred'' indicates the classification result by PCNN. ``1'' and ``0'' refer to self-contradiction and non-self-contradiction, respectively. The \textbf{bold} texts are manually highlighted to indicate the possible contradiction parts in the sentence identified by PCNN.}
\label{tab:casestd}
\resizebox{1.0\textwidth}{!}{%
\begin{tabularx}{\textwidth}{c|X}
\hline
 & \underline{GT=1 and Pred=1} \\ \hline
Article 1 & Title: \textit{Athanasius II of Constantinople} \\ \hline
s1 & He supposedly served from \textbf{1450 to 1453} \\ \hline
s2 & Athanasius II of Constantinople In office \textbf{1451 -- 1453} \\ \hline
Article 2 & Title: \textit{The Silent Scream (1979 film)} \\ \hline
s1 & The film was released theatrically by American Cinema Releasing in limited theaters in \textbf{November 23, 1979} in Victor, Texas, and in \textbf{January 30, 1980} in Bismarck, North Dakota. \\ \hline
s2 & The film is released \textbf{1980/1/18} \\ \hline\hline
 & \underline{GT=1 and Pred=0} \\ \hline
Article 1 & Title: \textit{Embassy of Cambodia in Washington, D.C.} \\ \hline
s1 & His Excellency Sounry Chum is the current \textbf{Cambodian Ambassador to the United States}, and was appointed to the role in 2018. \\ \hline
s2 & Royal \textbf{Embassy of Cambodia in Washington, D.C} is the diplomatic mission of the Kingdom. \\ \hline
Article 2 & Title: \textit{Arthur Desmond} \\ \hline
s1 & As with most aspects of Arthur Desmond's life, his \textbf{birth statistics are problematic}, and Arthur Desmond spent his adult life \textbf{concealing his origins as well as his identity.} \\ \hline
s2 & \textbf{Whatever his real origins}, the first concrete evidence of Arthur Desmond’s life comes when he stood for parliament in Hawke’s Bay. \\ \hline\hline
 & \underline{GT=0 and Pred=1} \\ \hline
Article 1 & Title: \textit{Calcio Fiorentino} \\ \hline
s1 & They try to \textbf{pin and force into submission as many players possible}. Once there are enough incapacitated players, the other teammates come and swoop up the ball and head to the goal. \\ \hline
s2 & It is also \textbf{prohibited for more than one player to attack an opponent}. Any violation leads to being expelled from the game. \\ \hline
Article 2 & Title: \textit{House of Terror (1960 film)} \\ \hline
s1 & Casimiro (Tin Tan), the night watchman at a wax museum of horrors, has \textbf{been napping more frequently} on the job because his boss, Professor Sebastian (Yerye Beirute). \\ \hline
s2 & As he \textbf{struggles to awareness}, the clouds outside part, the full moon shines on his face through a window, and the resurrected corpse transforms into a werewolf. \\ \hline
\end{tabularx}%
}
\end{table*}

\textbf{Imbalance Analysis.}
We further aim to investigate whether the proposed PCNN can survive from the imbalance between self-contradiction (P) and non-self-contradiction articles (N). By fixing the training percentage 80\%, we change the imbalance ratios, and report the performance comparison in Table~\ref{tab:imbalanced}. The results bring few findings. First, although a more imbalanced setting (e.g., P:N=10\%:90\%) causes performance drop across all methods, PCNN is still able to consistently outperform the competing models. This shows the superiority of PCNN when facing class imbalance. Second, although the Precision scores are quite low in the imbalanced setting of P:N=10\%:90\%, the corresponding Recall scores can be maintained as similar as those in P:N=30\%:70\% and P:N=50\%:50\%. The results indicate when regarding detecting all self-contradiction articles, the capability of PCNN is still effective in the imbalanced setting since the number of self-contradiction articles is small. 




\subsection{Case Study}
To validate whether PCNN can comprehensively highlight the most contradictory sentence pairs in an article, we conduct the case study. We consider three scenarios: both the ground-truth (\textbf{GT}) and PCNN prediction (\textbf{Pred}) are self-contradictory, and the inconsistency between the ground-truth and PCNN prediction. By reporting two articles for each scenario, and exhibiting the most contradictory sentence pair highlighted by PCNN, we display the results of case studies in Table~\ref{tab:casestd}. For the scenario ``GT=1 and Pred=1'', we can find that PCNN can nicely identify the pair of sentences that contradict one another. For the scenario ``GT=1 and Pred=0'', PCNN predicts Article 1 as non-self-contradiction since it considers ``Cambodian Ambassador to the United States'' in s1 is semantically consistent with ``Embassy of Cambodia in Washington, D.C''. In Article 2, PCNN also feels ``birth statistics are problematic'' in s1 does not contradict with ``Whatever his real origins'' in s2. For the scenario ``GT=0 and Pred=1'', in Article 1, PCNN is confused by ``submissions as many players possible'' in s1 and ``prohibited for more than one player'' is s2, i.e., PCNN considers they are contradictory. Besides, PCNN thinks Article 2 is self-contradiction because ``napping more frequently'' contradicts with ``struggles to awareness''. In summary, although PCNN can make incorrect classification, the sentence pairs highlighted by PCNN can reasonably explain the prediction results. That said, PCNN can identify inconsistent parts between sentences even the article is non-self-contradiction.

\section{Discussion}
\label{sec:discuss}
We discuss issues regarding the proposed self-contradiction detection model PCNN. 
The extensions of PCNN can be summarized in the following two points.
\begin{itemize}
\item \textbf{Pre-Training PCL.} The current PCNN utilizes the datasets on textual entailment, i.e., SNLI and MNLI, to pre-train the pairwise contradiction learning (PCL) module. We think using more datasets with the contradiction label to pre-train PCL can cover more diverse aspects on contradiction, and thus can benefit the fine-tuning for self-contradiction article detection. Datasets with contradictory sentence pairs employed by the relevant studies, as listed in Table~\ref{tab:relatedtab}, can be candidates to extend the coverage of predictable self-contradiction topics.
\item \textbf{Article-level Contradiction.} PCNN is now devised to model whether any sentence pairs are contradictory, i.e., sentence-level contradiction. In fact, PCNN can serve as a framework to allow the detection of article-level contradiction. To be specific, given a corpus containing a number of articles, PCNN can be moderately extended to detect whether two articles contradict with each other. We can change ``sentence pairs'' to ``article pairs'' in the encoders of PCNN.
\end{itemize}

There are two limitations in the current PCNN model. One is about the \textit{subjectivity} of self-contradiction, and the other is being not able to deal with \textit{long documents}.
\begin{itemize}
\item \textbf{Subjectivity.} Determining whether an article contains contradiction elements is subjective~\cite{dis3,dis2}. Although we have experienced and high-quality Wikipedia editors to annotate the ``Self-Contradictory Template'' to articles, it is unknown about how the subjectivity of editors affects the annotated self-contradiction labels. While our PCNN highly relies on the self-contradiction labels to fine-tune the PCL network, the degree of annotation subjectivity can to some extent influences how PCNN models contradiction and the detection performance. However, PCNN does not deal with the subjectivity issue. We think that mitigating the subjectivity on annotated self-contradictions in the future extension of PCNN will help improve the performance.
\item \textbf{Long Documents.} In the pairwise contradiction learning of PCNN, we only model sentences within a paragraph. However, for long documents, the contradiction could appear in sentences between paragraphs. PCNN can fail to detect self-contradiction for long articles. To deal with such an issue, one may resort to long-document transformer~\cite{dis4} to process long sequences of sentences and paragraphs in the encoding parts of PCNN.
\end{itemize}

PCNN can be directly utilized for two applications that concern the contradiction or inconsistency between texts. The first is to detect \textit{incongruence} between news headline and its body text~\cite{dis1}. This task can be performed if we put the news headline and body text together as an article, which can be directly fed into PCNN for incongruence detection. The second is to detect whether an article \textit{contradicts with other articles} in Wikipedia. We can collect the Wikipedia articles that contradict with other articles through the ``Contradicts others'' template\footnote{\url{https://en.wikipedia.org/wiki/Template:Contradicts_others}}\footnote{\url{https://en.wikipedia.org/wiki/Category:Articles_contradicting_other_articles}}. By merging every pair of articles into a long article and treating such a long article as the input of PCNN, we can classify whether it contains contradiction.

\section{Conclusions}
\label{sec:conc}
In this paper we defined and developed a solution for the task of finding self-contradiction articles in Wikipedia. We create the first dataset, WikiContradiction for this task. We also present the first model, Pairwise Contradiction Neural Network (PCNN). The most essential component of PCNN is pairwise contradiction learning, which is pre-trained on SNLI and MNLI datasets and fine-tuned on our dataset. 
The empirical results exhibit promising performance of PCNN. The case study also shows the model explainability of PCNN. We believe this work can be a pioneer study on self-contradiction article detection. The compiled WikiContradiction dataset can be a training resource for improving the quality of Wikipedia articles, and further contribute to fact-checking and claim verification. Experimental results also point out where future work can improve, including sentence representation learning, pairwise contradiction reasoning, and finer-grained selection of sentence pairs.


\section*{Acknowledgment}
This work is supported by Ministry of Science and Technology (MOST) of Taiwan under grants 110-2221-E-006-136-MY3, 110-2221-E-006-001, and 110-2634-F-002-051.

\small
\bibliography{wiki}
\bibliographystyle{IEEEtranN}



%



\end{document}